\PassOptionsToPackage{unicode}{hyperref}
\PassOptionsToPackage{hyphens}{url}
\documentclass[
  11pt,
]{article}
\usepackage{amsmath,amssymb}
\usepackage{iftex}
\ifPDFTeX
  \usepackage[T1]{fontenc}
  \usepackage[utf8]{inputenc}
  \usepackage{textcomp} % provide euro and other symbols
\else % if luatex or xetex
  \usepackage{unicode-math} % this also loads fontspec
  \defaultfontfeatures{Scale=MatchLowercase}
  \defaultfontfeatures[\rmfamily]{Ligatures=TeX,Scale=1}
\fi
\usepackage{lmodern}
\ifPDFTeX\else
\fi
\IfFileExists{upquote.sty}{\usepackage{upquote}}{}
\IfFileExists{microtype.sty}{% use microtype if available
  \usepackage[]{microtype}
  \UseMicrotypeSet[protrusion]{basicmath} % disable protrusion for tt fonts
}{}
\makeatletter
\@ifundefined{KOMAClassName}{% if non-KOMA class
  \IfFileExists{parskip.sty}{%
    \usepackage{parskip}
  }{% else
    \setlength{\parindent}{0pt}
    \setlength{\parskip}{6pt plus 2pt minus 1pt}}
}{% if KOMA class
  \KOMAoptions{parskip=half}}
\makeatother
\usepackage{xcolor}
\usepackage{longtable,booktabs,array}
\usepackage{calc} % for calculating minipage widths
\usepackage{etoolbox}
\makeatletter
\patchcmd\longtable{\par}{\if@noskipsec\mbox{}\fi\par}{}{}
\makeatother
\IfFileExists{footnotehyper.sty}{\usepackage{footnotehyper}}{\usepackage{footnote}}
\makesavenoteenv{longtable}
\usepackage{graphicx}
\makeatletter
\def\maxwidth{\ifdim\Gin@nat@width>\linewidth\linewidth\else\Gin@nat@width\fi}
\def\maxheight{\ifdim\Gin@nat@height>\textheight\textheight\else\Gin@nat@height\fi}
\makeatother
\setkeys{Gin}{width=\maxwidth,height=\maxheight,keepaspectratio}
\makeatletter
\def\fps@figure{htbp}
\makeatother
\ifLuaTeX
  \usepackage{luacolor}
  \usepackage[soul]{lua-ul}
\else
  \providecommand{\ul}[1]{\underline{#1}}
\fi
\usepackage[letterpaper,margin=1in]{geometry}
\usepackage{mathptmx}
\usepackage{amssymb}
\usepackage{newunicodechar}
\usepackage{setspace}
\usepackage{float}
\usepackage[font=small,labelfont=bf]{caption}
\usepackage{microtype}
\newunicodechar{β}{\ensuremath{\beta}}\newunicodechar{α}{\ensuremath{\alpha}}\newunicodechar{γ}{\ensuremath{\gamma}}
\newunicodechar{δ}{\ensuremath{\delta}}\newunicodechar{ε}{\ensuremath{\varepsilon}}\newunicodechar{θ}{\ensuremath{\theta}}
\newunicodechar{λ}{\ensuremath{\lambda}}\newunicodechar{μ}{\ensuremath{\mu}}\newunicodechar{σ}{\ensuremath{\sigma}}
\newunicodechar{τ}{\ensuremath{\tau}}\newunicodechar{φ}{\ensuremath{\phi}}\newunicodechar{ω}{\ensuremath{\omega}}
\newunicodechar{π}{\ensuremath{\pi}}\newunicodechar{η}{\ensuremath{\eta}}\newunicodechar{ρ}{\ensuremath{\rho}}
\newunicodechar{Δ}{\ensuremath{\Delta}}\newunicodechar{Σ}{\ensuremath{\Sigma}}\newunicodechar{Ω}{\ensuremath{\Omega}}
\newunicodechar{κ}{\ensuremath{\kappa}}\newunicodechar{ξ}{\ensuremath{\xi}}\newunicodechar{ψ}{\ensuremath{\psi}}\newunicodechar{χ}{\ensuremath{\chi}}\newunicodechar{ϕ}{\ensuremath{\phi}}
\newunicodechar{≈}{\ensuremath{\approx}}\newunicodechar{≤}{\ensuremath{\leq}}\newunicodechar{≥}{\ensuremath{\geq}}
\newunicodechar{±}{\ensuremath{\pm}}\newunicodechar{×}{\ensuremath{\times}}\newunicodechar{→}{\ensuremath{\rightarrow}}
\newunicodechar{∈}{\ensuremath{\in}}\newunicodechar{∼}{\ensuremath{\sim}}\newunicodechar{−}{\ensuremath{-}}\newunicodechar{°}{\ensuremath{^\circ}}
\newunicodechar{µ}{\ensuremath{\mu}}\newunicodechar{·}{\ensuremath{\cdot}}\newunicodechar{∞}{\ensuremath{\infty}}\newunicodechar{√}{\ensuremath{\surd}}
\newunicodechar{‑}{-}\newunicodechar{​}{}
\usepackage[section]{placeins}

\AtBeginDocument{\setlength{\LTleft}{0pt plus 1fill}\setlength{\LTright}{0pt plus 1fill}}
\ifLuaTeX
  \usepackage{selnolig}  % disable illegal ligatures
\fi
\IfFileExists{bookmark.sty}{\usepackage{bookmark}}{\usepackage{hyperref}}
\IfFileExists{xurl.sty}{\usepackage{xurl}}{} % add URL line breaks if available
\hypersetup{
  hidelinks,
  pdfcreator={LaTeX via pandoc}}

\author{}
\date{}

\begin{document}
\begin{center}
{\Large\bfseries Human Predictive Control of Welding Robot through Generative AI Enhanced Human-Robot Collaboration\par}\vspace{8pt}
{\small Yue Cao\textsuperscript{1}, Ninshu Ma\textsuperscript{2}, Qiang Ye\textsuperscript{3}, YuMing Zhang\textsuperscript{1*}\par\vspace{2pt}
1: Department of Electrical and Computer Engineering and Institute for Sustainable Manufacturing, University of Kentucky, Lexington, KY 40506, USA\par\vspace{2pt}
2. Welding and Joining Research Institute, The University of Osaka, Japan\par\vspace{2pt}
3: Department of Mathematics, University of Kentucky, Lexington, KY 40506, USA\par\vspace{2pt}
\begin{quote}
*: \href{mailto:yuming.zhang@uky.edu}{\ul{yuming.zhang@uky.edu}}
\end{quote}\par}
\end{center}

Abstract: Human-robot collaboration (HRC) provides a transformative approach to robotize complex manufacturing processes like arc welding, which requires enabling adaptive human control of the robotic process first and then extracting human intelligence for autonomy. However, an effective HRC welding system must address both physical execution through easy teleoperating ability and the enhancement of the operator\textquotesingle s adaptive decision-making capabilities. In this work, we developed an easily operable HRC system for Virtual Reality (VR)--based robotic teleoperation, enabling welding control with reduced physical and cognitive loads on the operator. Specifically, the system allows the human operator to remotely adjust welding torch motion via a handheld handler, while accessing an immersive VR environment that displays the multi-source real-time information, overcoming the danger and limitation of direct on-site operation. At its core is a strategically designed generative model, Hybrid Latent Variational Autoencoder (HL-VAE), that decouples structural and stochastic information on weld pool images in the latent space, enabling precise long-term prediction of future imaged weld pool evolution. Thus, the predictive capability of skilled welders to anticipate weld pool evolution can be replicated by the system, reducing cognitive burden and supporting proactive decision-making. Comparative experiments verify that the developed system enables rapid and precise adjustments. This work exemplifies a human-centric HRC paradigm that unites human expertise, robotic execution, and computational intelligence for intelligent manufacturing.


\hypertarget{introduction}{%
\section{Introduction}\label{introduction}}

Manufacturing is undergoing rapid transformation, becoming increasingly efficient, agile, and productive. Robotization is a major driver of this shift, and welding has long been at the forefront. According to the International Federation of Robotics, ``50\% of the world's robots are used for welding'' {[}1{]}. Of these, ``33\% are used for spot welding, 16\% for arc welding, and 1\% for other types of welding operations'' {[}1{]}. This distribution is unsurprising: spot welding primarily requires electrode positioning, which demands little adaptation from robots. In contrast, arc welding requires substantial adaptability---something human welders naturally provide---making robotization far more challenging due to strong process-specific dependencies. This challenge is especially evident in Gas Tungsten Arc Welding (GTAW), where a non-consumable electrode enables precise manipulation of arc heat for adaptive control. Such controllability makes GTAW highly desirable for precision joining in critical applications {[}2{]}, but its complexity also makes real-time robotic control particularly difficult.

A central challenge lies in the dynamic evolution of the weld pool, which ultimately determines weld quality. The relationship between manipulated welding parameters (control variables) and the resulting weld pool (outputs) is highly sensitive to manufacturing conditions such as material properties, joint geometry, welding position, and environment---factors that frequently deviate from nominal values {[}3{]}. Because the weld pool evolves dynamically, control variables must be adjusted in real time. Yet uncertainties introduced by such deviations make effective control difficult. Conventional adaptive control approaches {[}4{]}, which must be tailored to specific applications, have not provided a scalable solution for robotization. Skilled human welders can meet these challenges, but their expertise requires extensive training and practice. Even then, consistency is not guaranteed, as welding environments are harsh and both physical and cognitive demands remain high.

Robotization can instead proceed through a two-step pathway: (1) enabling the robot to be controlled by a human for adaptive response, and (2) deploying the learned human adaptation model to the robot. This is more efficient than directly transferring a human response model from manual welding to robotic welding {[}5{]}. Manual operation allows direct observation of how humans respond to process evolution but requires robotic operation to replicate manual operation exactly. Critically, this also requires welders to carry sensors to capture process evolution and manipulation data, or a separate robot to track human operation. Either approach complicates implementation and limits scalability.

Allowing humans to operate the robot directly increases scalability by reducing the skill level required of welders. Human welding skills can be divided into two categories: (i) the ability to decide what manipulation (control) is needed, and (ii) the ability to execute the decided manipulation. The real-time coupling of these two abilities makes manual welding particularly demanding. By having humans operate through a robot, most execution-related skills are no longer required. Since robot and torch operations differ, effective execution may be achievable by less-trained operators under suitable conditions. This work realizes such an approach through enhanced human--robot collaboration (HRC), as detailed in Section 3.

The ability to decide what manipulation is needed remains critical, yet it is application dependent and even skilled welders require practice to develop it. The availability of qualified welders thus limits scalability. We hypothesize that, under suitable conditions, less-trained operators can effectively decide necessary manipulations in response to process evolution. Because execution burdens are reduced, the challenge of decision-making can be partially alleviated. To further reduce training and practice demands, this work proposes providing humans with predictive information on future process evolution, enabling them to anticipate and decide on required manipulations. For skilled welders, this also reduces the practice needed to gain application-specific response patterns. We realize this through a generative model that predicts future outputs based on process history and intended manipulations. While application dependent, the model can be trained on data from the Human--Robot Collaborative Welding System (HRC-WS). As this requires predicting human visual observations of the weld pool---a complex task---we propose leveraging generative artificial intelligence (AI), as detailed in Section 4.

Taken together, this work proposes and demonstrates an effective road map to robotize arc welding. Consider a robotic welding system that is already equipped with sensors but lacks an adaptation model to respond to observed process evolution. The proposed road map consists of: (1) integrating a human--robot interface (HRI) that allows a human to observe and control the dynamically evolving process, and (2) transferring the learned human response model to the robot for full autonomy. Because the addition of an HRI can be standardized and its cost is negligible relative to the overall robotic system, this approach does not impose significant additional burden. Importantly, execution skills can be practiced without welding, and data for training generative prediction models can be generated through programmed variations in control parameters. This enables humans to prepare for real-time adaptive response with minimal welding trials. Consequently, humans can achieve effective adaptive control with significantly reduced training and practice.

This paper is organized as follows: Section II reviews related work. Section III introduces the configuration of the developed HRC welding system. Section IV discusses the principles of image processing. Section V presents experimental verification results. Finally, Section VI concludes the paper and outlines directions for future work

\hypertarget{related-work}{%
\section{Related Work}\label{related-work}}

This section reviews the advancements of HRC for welding and other manufacturing applications. HRC systems in manufacturing incorporate functionalities such as human sensors, signal processing, interface control, and information presentation {[}6{]}. They allow safe physical interaction and improved production performance by matching human inimitable skills with machine strengths {[}7{]}. However, applying these systems to welding remains a complex challenge due to the highly process-specific nature of welding operations. In this section, representative HRC approaches for welding are reviewed.

\hypertarget{hrc-for-welding-path-planning}{%
\subsection{HRC for Welding Path Planning}\label{hrc-for-welding-path-planning}}

As the prior step of robotic welding tasks, path planning has been addressed by HRC to adapt to varying welding positions and complex workpiece geometry. Human-guided depth camera scanning can generate a 6-DOF welding path for the robot {[}8{]}. Offline programming methods further allow operators to trace metal joints and visualize virtual weld seams through augmented reality (AR), with captured motions converted into robot code {[}9{]}. In {[}10{]}, a robot-assisted welding solution enables welders to intuitively program paths via a control panel by setting key points and parameters. Vision-based systems combining deep learning joint detection and laser scanning have also been proposed, with human feedback used to correct detection failures {[}11{]}.

\hypertarget{hrc-welding-teleoperation}{%
\subsection{HRC Welding Teleoperation}\label{hrc-welding-teleoperation}}

Beyond path planning, most HRC studies focus on teleoperated robotic welding to mitigate the physical limitations and hazards of manual welding. Ergonomic hand-motion control has been enhanced using extrapolation algorithms to enable continuous operation and reduce fatigue {[}12{]}. Haptic-enabled systems provide mechanical impedance feedback to convey collisions between the robot and workpiece {[}13{]}, while Kalman filtering with disturbance observers has been used to restore accurate, low-latency hand motion in remote arc welding {[}14{]}. Virtual reality has also been widely adopted, blending real welding views with robot digital twins to create immersive and informative workspaces {[}15{]}.

\hypertarget{integrated-hrc-welding-systems}{%
\subsection{Integrated HRC Welding Systems}\label{integrated-hrc-welding-systems}}

Integrating multiple technologies, HRC welding systems address diverse application scenarios. A dual-robot HRC system for Double Electrode Gas Metal Arc Welding (DE-GMAW) enables stable, real-time electrode position adjustment within millimeter-scale regions {[}16{]}. Object detection networks combined with Gaussian Process Regression have been used to enhance human observation and model human response intelligence for electrode adjustment {[}17{]}. Weld-pool morphology acquisition systems have also been developed, fusing current, voltage, and image data to reconstruct and visualize 3D weld pool geometry in VR for precise teleoperation {[}18{]}. The Versatile Remediation Module further demonstrates an integrated robotic platform for remote inspection and GTAW-based weld repair of nuclear storage canisters under full operator supervision {[}19{]}.

\hypertarget{learning-strategies-in-welding-hrc}{%
\subsection{Learning Strategies in Welding HRC}\label{learning-strategies-in-welding-hrc}}

In addition to enhancing human observation and operation, HRC systems increasingly incorporate learning strategies to model human intelligence in welding and related manufacturing processes. For weld pool control, machine algorithms combined with learned welder-response models use numerical metrics from structured-laser sensing to achieve robust welding-speed regulation {[}20{]}, while CNN--LSTM models trained on human demonstrations directly translate visual weld-pool feedback into speed adjustments {[}21{]}. At the task-planning level, motion and process parameters from skilled human operations are encoded into reusable skill libraries, enabling robots to reproduce skilled-welder performance {[}22{]}. To support remote skill development, robotic systems capture expert motion patterns and provide VR-based virtual tutoring to guide trainees {[}23{]}. Beyond welding, similar learning-from-demonstration strategies have been applied to shipbuilding linear heating via reinforcement learning {[}24{]} and to machining tasks using Gaussian process--based controllers for adaptive force and stiffness regulation {[}25{]}.

\hypertarget{gap-in-arc-welding-process-control}{%
\subsection{Gap in Arc Welding Process Control}\label{gap-in-arc-welding-process-control}}

However, the main challenge in arc welding is process control. The difficulty lies in that what can be conveniently and reliably measured (observables) is typically not exactly the same as the output the process control aims to achieve (weld penetration). Existing studies commonly employ deep learning models trained via supervised learning to extract manually defined numerical feedback signals from these observables {[}26{]}. Such feedback design is inherently limited by the designer's expertise and prior assumptions, and lacks the semantic interpretation of weld pool behavior possessed by skilled welders. Furthermore, although advanced strategies such as Model Predictive Control (MPC) {[}27{]}, adaptive control {[}28{]}, and intelligent control {[}29{]} leverage this feedback for closed-loop regulation, their efficacy is often limited by the use of simplified numerical feedback that fails to capture the visual essence under complex and extreme conditions.

The involvement of humans mitigates all these challenges/difficulties naturally. However, high training costs, physical limitations in strength, vision, and endurance, along with environmental hazards, limit the scalability of manual welding. HRC can reduce the demands on humans in training and concentration so that control is further assured in a systematic way when humans are provided with enhanced observables and eased operations that allow them to focus on their adaptation/reaction to the dynamically fluctuating process. For example, a digital twin framework empowered by deep learning enabled real-time monitoring and penetration control by estimating weld joint growth from multi-sensor data {[}30{]}, and a subsequent system further improved operational comfort in GTAW by mirroring human torch manipulation to robot execution, creating a VR-based digital twin, and incorporating data-driven behavior analysis {[}31{]}.

Such behavior resembles model predictive control {[}3{]} but is distinctly enriched by the human semantic understanding of weld pool images and process dynamics, an ability not yet fully replicated by algorithms. To reduce reliance on this experiential skill, it is essential to provide operators with predictive weld pool images that are both physically consistent and visually interpretable. Embedding such predictive modeling into HRC systems can lower cognitive demands on the operator, reduce the experiential threshold for effective welding process control, and simplify decision-making. However, achieving this capability requires a physically consistent, high-fidelity image generation model and seamless integration with HRC systems, which presents significant technical challenges.

\hypertarget{enhanced-human-robot-collaboration-system}{%
\section{Enhanced Human-Robot Collaboration System}\label{enhanced-human-robot-collaboration-system}}

Figure 1 illustrates the components of the system for enhanced HRC-WS and their connections. The system enables human-in-the-loop control of a GTAW welding process, with a VR environment presenting real-time feedback and generative model predictions.

\begin{figure}[!htbp]
\centering
\includegraphics[width=4.06087in,height=3.2208in]{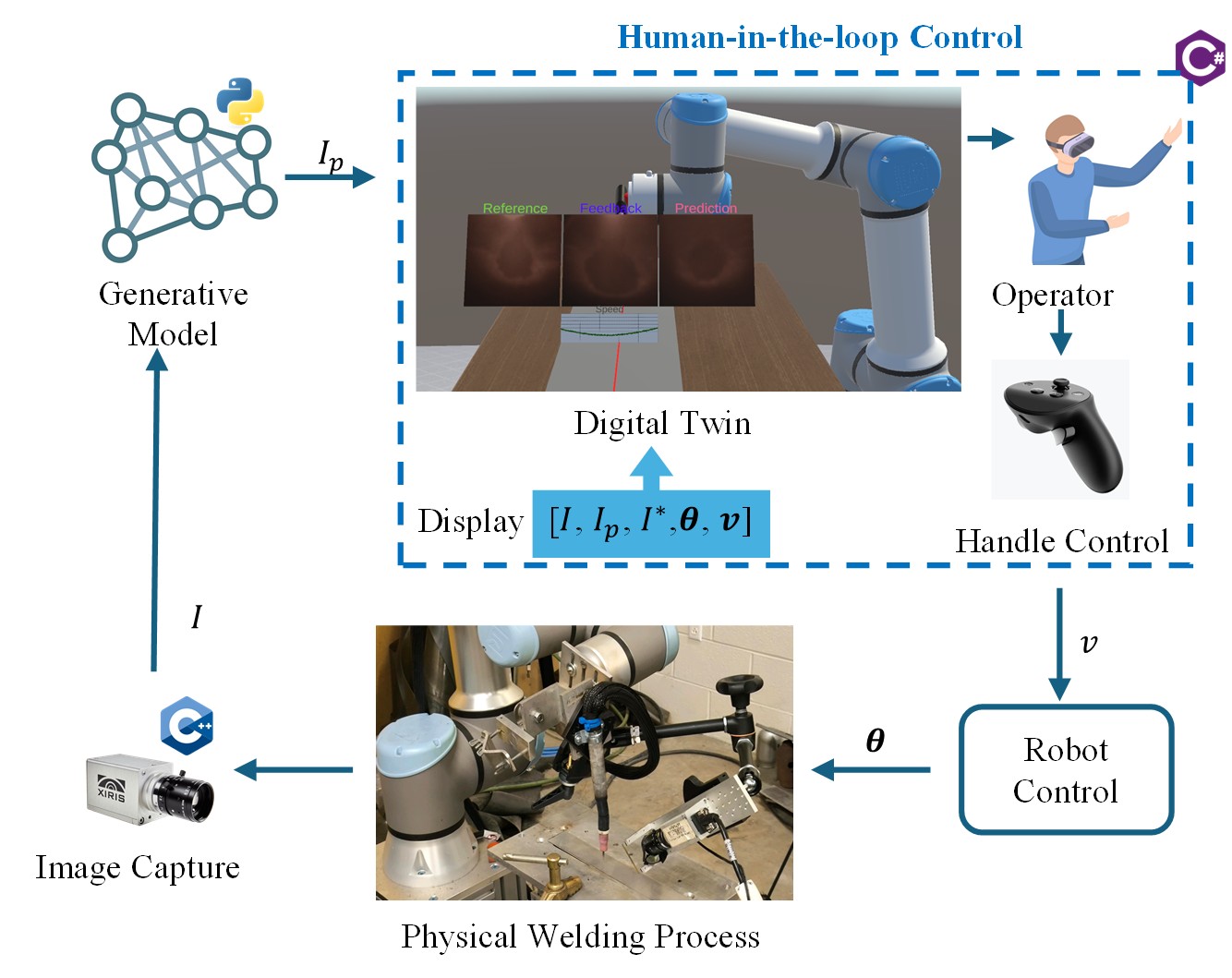}
\caption*{Figure 1. System configuration. \(I\): process raw image; \(I^{*}\): referred image; \(I_{P}\): predictive images; \(\mathbf{v}\): welding speed; \(\mathbf{\theta}:\) joint angle command.}
\end{figure}

Robotic welding is realized with a welding system (including welding power source, torch, gas cylinder, etc.), and a UR10e robot for traversal speed adjustment. The Xiris weld camera XVC-1100, a digital weld camera intended for open arc welding, is mounted on the robot to capture the images of arc and weld pool. A 650\,nm band-pass optical filter is attached to the lens to filter the strong arc light. The welding process is pulsed GTAW, using DCEN polarity, where the operator adjusts welding speed to regulate the front-side weld pool status through the virtual welding environment. A pulsed current is applied, with the peak current providing sufficient penetration and the base current maintaining arc stability. The peak and base currents are set to 100\,A and 20\,A, respectively, with a pulse frequency of 10\,Hz and a 50\% duty cycle. Weld pool images are recorded during the base phase (one frame per cycle) to ensure clear visualization of pool boundaries. The workpiece material is stainless steel 304L with a thickness of 1.85\,mm. The welding speed ranges from 2.5 to 5.5\,mm/s and is regulated remotely by the operator through the system.

VR based operation is achieved via a Meta Quest Pro VR headset and Touch Pro controllers. The headset offers 1800×1920 resolution per eye at a 90\,Hz refresh rate for a vivid experience, and the controllers provide 6-DOF motion tracking. Within the VR environment, the digital twin (DT) acts as an information synthesis platform, combining data from various sources for human perception. Primarily, the human can access the real-time weld pool stream from camera capture \(I\) as process feedback, a referred weld pool view \(I^{*}\) as control target, and generative model prediction \(I\textsubscript{p}\). The generative model utilizes the current feedback image, stored historical information, and the operator's speed command to generate a near-future weld pool image \(I\textsubscript{p}\), assuming the speed command is applied and fixed in the future. By presenting both predictive and real-time views, the interface allows operators to evaluate and refine control actions proactively. The welding speed command \(\mathbf{v}\), and a virtual robot are also presented.

Human operation is integrated through a control logic that maps 6-DOF handle motion to robot execution (Figure 2). The right controller captures vertical hand displacement, where 100 mm corresponds to a 1 mm/s speed change, and converts it into a filtered speed command \(\mathbf{v}\). The robot control program then updates the desired end-effector pose \(P^{R} \leftarrow P^{R} + {\mathrm{\Delta}t\lbrack v_{x},v_{y},v_{z},0,\ 0,\ 0\rbrack}^{T}\) to generate joint angles \ensuremath{\theta}. During welding, only the directional velocity \(v_{x}\), is allowed, with the other two components set to zero. During repositioning, all three velocity components are enabled, allowing the operator to adjust torch position in a small range for seam tracking. The virtual robot synchronizes in real time by copying the joint angles \ensuremath{\theta} from the physical robot.

\begin{figure}[!htbp]
\centering
\includegraphics[width=3.49361in,height=1.43564in]{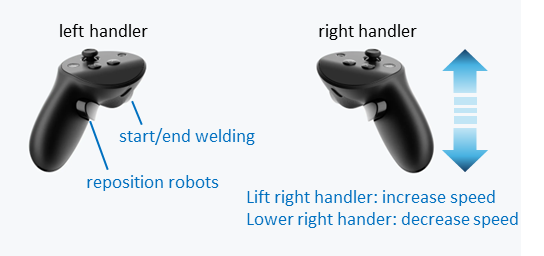}
\caption*{Figure 2. Handler input logic.}
\end{figure}

In summary, the HRC-WS achieves smooth integration of robotic execution, VR-based perception, and generative model--enhanced process state forecasting, thereby augmenting human intelligence in GTAW process control. Teleoperation of the welding robot and immersive virtual multi-source information display alleviate the physical demands of on-site welding, while predictive generation of future weld pool views reduces the cognitive skill required for adaptive process regulation. Together, these innovations shift the operator's role from skilled manual execution toward higher-level decision-making, establishing a human-robot collaboration paradigm that leverages the complementary strengths of humans, machines, and computational intelligence. A demonstration of the system is available at \url{https://github.com/YueCaoUKY/HL-VAE-Welding/blob/main/DemoVideo.mp4}.

\hypertarget{generative-weld-pool-prediction}{%
\section{Generative Weld Pool Prediction}\label{generative-weld-pool-prediction}}

\hypertarget{latent-representation-of-weld-pool-views}{%
\subsection{Latent Representation of Weld Pool Views}\label{latent-representation-of-weld-pool-views}}

The key innovation of the HRC system lies in its ability to provide predictive weld pool views, enabling the operator to anticipate process evolution rather than reactively responding. This is achieved by employing a generative model for the synthesis of weld pool image dynamics, which arises from the nature of welding physics. However, efficient representation of complex manufacturing data, as the core of the generative model, is critical to reduce computational burden and enhance predictive accuracy.

In this work, we propose a Hybrid Latent Variational Autoencoder (HL-VAE), a novel latent representation framework for data in manufacturing scenarios with structural information and yet noise and interference such as GTAW. Although the weld pool view \(x\) appears visually complex, its underlying evolution is governed primarily by a limited set of changing welding parameters, through underlying physics such as heat transfer and fluid dynamics, causing patterned changes in a low-dimensional manifold. It is shown that three key weld pool geometric characteristics, width, length, and convexity are statistically sufficient to predict the process output, weld penetration {[}32{]}. Dimensionless analysis also demonstrates that key aspects of weld pool appearance can be described using a few dimensionless numbers, which are combinations of process parameters or physical properties computed to eliminate units {[}33, 34{]}. Additionally, stochasticity exists in the physical processes. This stems from unpredictable factors such as turbulent molten metal flow and arc plasma fluctuations arising from process variations and material inconsistencies, causing stochastic visual cues such as irregular weld pool boundaries and arc radiation interference.

To systematically capture both components of weld pool appearance, HL-VAE develops a unified latent representation framework that parallels this physical decomposition by modeling the image \(x\) into \(x = x_{s} + x_{r}\), where (i) \(x_{s}\) contains structural content which is continuous in a low-dimensional manifold and (ii) \(x_{r}\) contains stochastic details which are discrete and diverse. Specifically, HL-VAE (Figure 3) employs two specialized branches to learn these distinct representations: a structural latent path and a stochastic residual path.

In the structural latent path, a Variational Autoencoder (VAE) is used to model the dominant weld pool geometry. Given an input image \(x\), the encoder \(E_{s}\) maps it to the parameters of a multivariate Gaussian distribution, producing the mean vector \(\mu\) and log-variance \(\log\sigma^{2}\). The decoder \(D_{s}\) then reconstructs a coarse weld pool image \(x_{s}\) from samples from the parameterized distribution. The variance \(\log\sigma^{2}\) absorbs noise for stable training, so only the mean vector \(\mu\) is retained during inference as the continuous structural latent representation.

The image residual, defined as \(x_{r} = x - x_{s}\), contains spatially localized and diverse high-frequency details, which are distinctive from the structural information. To encode such details, the stochastic residual path adopts a Residual Vector Quantization (RVQ), where the encoder \(E_{r}\) maps \(r\) into a dense feature tensor \(Z \in \mathbb{R}^{H_{z} \times W_{z} \times D_{z}}\). Here, each spatial location \((h,w)\) in \(Z\) contains a local feature vector. To efficiently represent the underlying structure of these localized features, we use a learned codebook \(\mathcal{C = \{}e(k) \in \mathbb{R}^{D_{z}}\}_{k = 1}^{K}\), where each embedding captures a reusable pattern common across many images.

\begin{figure}[!htbp]
\centering
\includegraphics[width=6in,height=4.42361in]{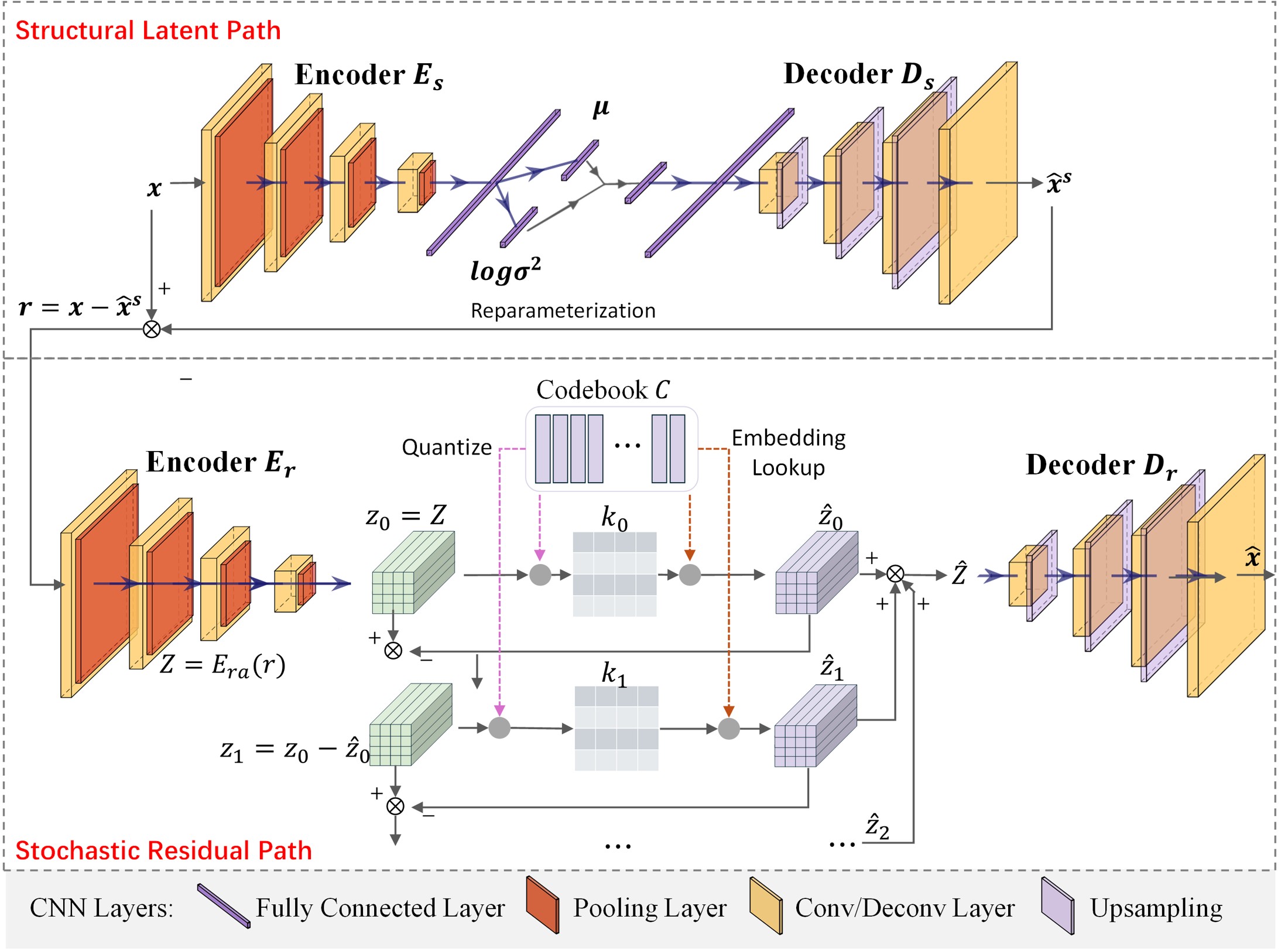}
\caption*{Figure 3. HL-VAE Structure.}
\end{figure}

At each location \((h,w)\), the quantization begins with the initial feature \(z_{0} = Z\), and the first-level approximation, \({\widehat{z}}_{0}\), is obtained by selecting the nearest embedding:

\[\begin{array}{r}
\begin{matrix}
k_{0}^{(h,w)} = {\arg\ \min}_{k \in \lbrack K\rbrack} \parallel z_{0}^{(h,w)} - e(k) \parallel_{2}^{2} \\
{\widehat{z}}_{0}^{(h,w)} = e\left( k_{0}^{(h,w)} \right)
\end{matrix}\tag{1}
\end{array}\]

The residual is then refined by subtracting the selected embedding:

\[\begin{array}{r}
z_{1} = z_{0} - {\widehat{z}}_{0}\tag{2}
\end{array}\]

This process is repeated for \(d = 0,\ldots,D - 1\), yielding a sequence of quantized approximations \({\widehat{z}}_{d}^{(h,w)}\), with each code index \(matrix\) \(k_{d} \in \mathbb{R}^{H_{z} \times W_{z}}\) selecting embeddings from \(\mathcal{C}\) that minimize the residual at current depth d. This coarse-to-fine quantization enables progressive reconstruction of fine-grained detail. The summing of reconstructions at each depth \({\widehat{z}}_{d}\) finally constructs the quantized feature tensor \(\widehat{Z}\). The residual decoder \(D_{r}\) takes \(\widehat{Z}\) as input to reconstruct the stochastic component. The full set of code indices forms a stacked discrete code map:

\[\begin{array}{r}
M = \lbrack k_{0},k_{1},\ldots,k_{D - 1}\rbrack \in \lbrack K\rbrack^{H_{z} \times W_{z} \times D}\tag{3}
\end{array}\]

Together, the pair \((\mu,M)\) forms the final latent representation of the image, where \(\mu\) captures the global structural manifold and \(M\) compactly encodes stochastic detail via discrete code indices.

In this work, the weld pool images (Figure 4) are RGB frames of size 128×128×3. In the structural latent path, the encoder \(E_{s}\) consists of four convolutional blocks with pooling layers, progressively reducing the resolution while increasing channel depth to yield an 8×8×256 feature tensor. This tensor is flattened and passed through fully connected layers to produce the mean \(\mu\) and log-variance \(log\sigma^{2}\) of a 5D multivariate Gaussian. To reconstruct the structural image \(x_{s}\), the decoder \(D_{s}\) mirrors the encoder using four upsampling--deconvolutional blocks. Herein, the continuous latent dimension is specifically set to 5, grounded in the assumption that weld pool geometry evolves along a low-dimensional manifold. Prior dimensionless analysis studies have experimentally and computationally verified that a few quantities are sufficient to describe pool shape changes {[}32, 33{]}. Additionally, given that the changing welding speed solely drives the weld pool evolution in our experimental setup (Figure 4), a dimension of 5 is sufficient to capture structural information while avoiding redundancy.

\begin{figure}[!htbp]
\centering
\includegraphics[width=4.14536in,height=2.992in]{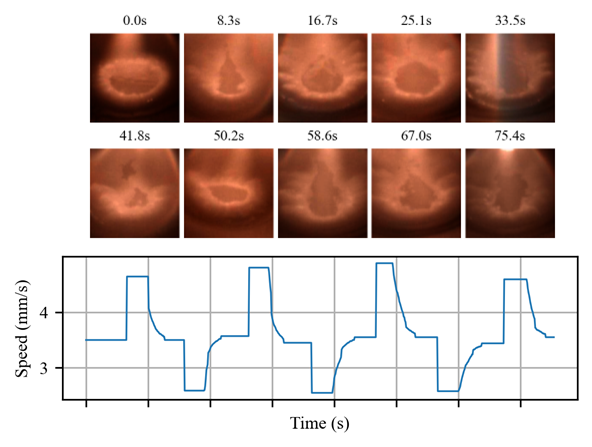}
\caption*{Figure 4. Experimental Data.}
\end{figure}

In the stochastic residual path, the image residual \(x_{r}\) (of size 128×128×3) is encoded by the encoder \(E_{r}\) into a compact feature tensor of size 8×8×64, where each 64-dimensional vector captures localized visual patterns. To discretize this representation, we employ a vector quantization codebook \(\mathcal{C \in}\mathbb{R}^{512 \times 64}\), consisting of \(K = 512\) learned embeddings, each of dimension \(D_{z} = 64\). Residual vector quantization is performed with depth \(D = 2\), meaning that each feature vector is iteratively approximated by a sum of two embeddings selected from \(\mathcal{C}\), progressively refining the representation. This results in a compact discrete index tensor \(M \in \lbrack K\rbrack^{8 \times 8 \times 2}\), storing two code indices per spatial location. The residual decoder \(D_{r}\) reconstructs the residual component \({\widehat{x}}_{r}\)from the quantized features, and the final image is obtained by combining both paths as \(x = x_{s} + {\widehat{x}}_{r}\).

To ensure the structural and stochastic paths are strictly decoupled, a sequential training strategy was employed. In Stage 1, the structural path is optimized under a β-TC-VAE framework {[}35{]} using the objective:

\[\begin{array}{r}
\begin{matrix}
\mathcal{L}_{s} = & \mathcal{L}_{MSE}\left( x,x_{s} \right) + \alpha\text{\,}MI(\mu;x) + \beta\text{\,}TC(\mu) \\
 & + \gamma\text{\,}KLD(\mu)
\end{matrix}\tag{4}
\end{array}\]

While the mean squared error (MSE) loss \(\mathcal{L}_{MSE}\) ensures reconstruction fidelity, the remaining terms regulate the 5D \(\mu\) for meaningful representation: Mutual Information (MI) (\(\alpha = 1.0\)) maximizes the information shared between the input and the latent code; Total Correlation (TC) (\(\beta = 6.0\)) suppresses statistical dependencies to disentangle \(\mu\) into independent physical components; and KL Divergence (KLD) \((\gamma = 1.0\)) regularizes the parameterized Gaussian space toward a standard normal distribution.

To validate that this training objective preserves physically meaningful information, Figure 5 illustrates the mapping between the learned structural latent components and the physical dimensions of the weld pool using data from one experiment. The weld pool width and area are extracted by a supervised-trained U-Net segmentation model {[}36{]} from one experiment, and the fitted values are obtained from five latent components using second-order polynomial regression. The fitted curves closely follow the measured geometric signals, with root mean squared error (RMSE) being 10.25 pixels for width and \(1.03 \times 10^{3}\) for area, indicating that the structural path encodes process-relevant characteristics rather than merely visually appealing features.

In Stage 2, the structural path is frozen, and the stochastic residual path is trained to reconstruct \(x_{r}\) using the combined loss:

\[\begin{array}{r}
\begin{matrix}
\mathcal{L}_{r} = & \mathcal{L}_{MSE}\left( x_{r},{\widehat{x}}_{r} \right) + \lambda\text{\,}\mathcal{L}_{perc}\left( x_{r},{\widehat{x}}_{r} \right) + \eta\text{\,}\mathcal{L}_{commit}\left( Z,\widehat{Z} \right)
\end{matrix}\tag{5}
\end{array}\]

This objective prioritizes high-frequency fidelity through a Perceptual Loss (\(\lambda = 0.1\)) which utilizes a pre-trained VGG network to extract and align high-level semantic features between \(x_{r}\) and its reconstruction \({\widehat{x}}_{r}\). The Commitment Loss (\(\eta = 0.25\)) stabilizes training by penalizing the distance between encoder features \(Z\) and their quantization \(\widehat{Z}\). Since quantization is non-differentiable, the codebook \(\mathcal{C}\) is updated via an Exponential Moving Average (EMA) strategy {[}37{]} with a decay of 0.95, ensuring embeddings adapt smoothly to the data distribution without the instabilities of direct backpropagation. Code and training details are at \url{https://github.com/YueCaoUKY/HL-VAE-Welding}.

\begin{figure}[!htbp]
\centering
\includegraphics[width=3.62135in,height=3.60417in]{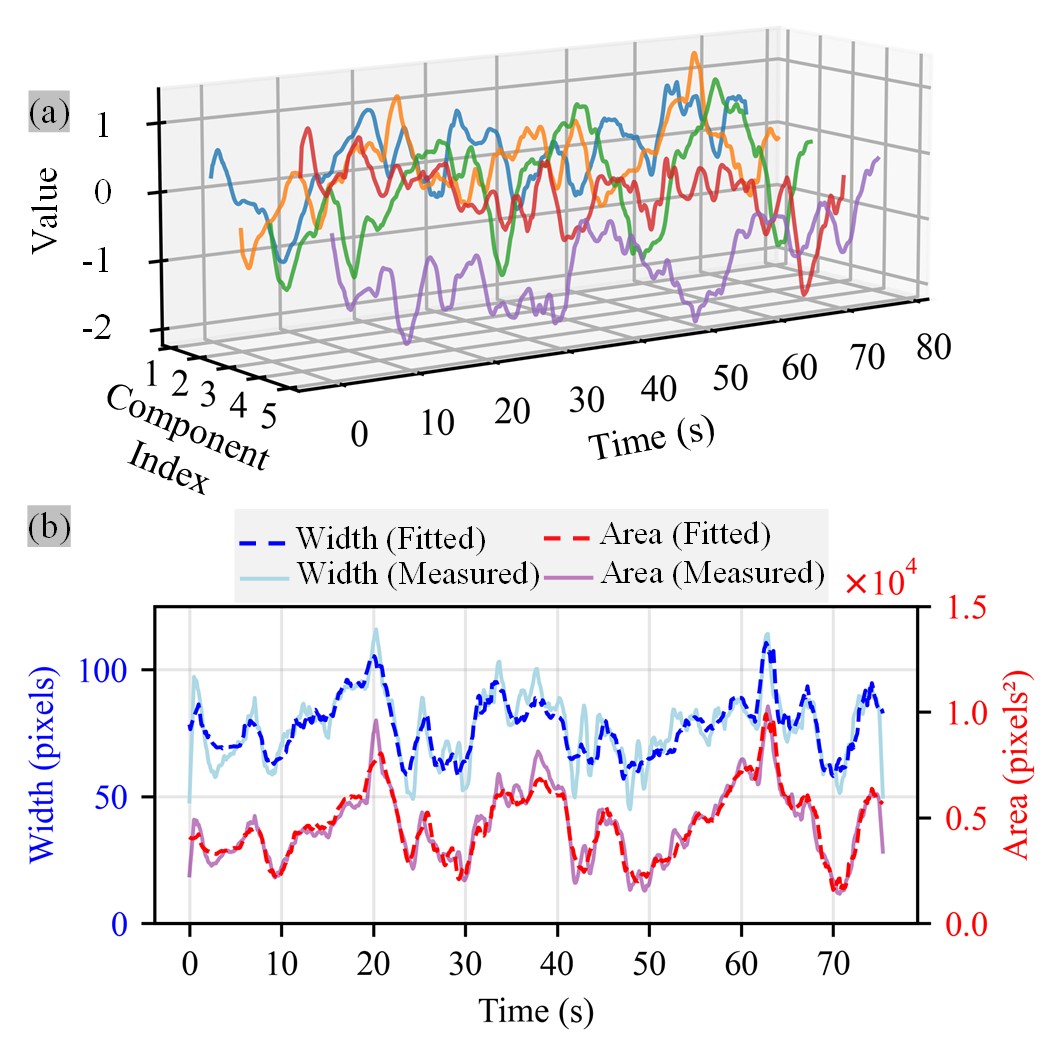}
\caption*{Figure 5. Physical Grounding of Structural Latent Representations. (a) Structural Latent Components; (b) Regression for Physical Dimensions.}
\end{figure}

Figure 6 compares the original weld pool image with HL-VAE reconstructions, including the structural output \(x_{s}\) from the continuous VAE path and the full reconstructions obtained at different residual depths. The reconstruction produced by the VAE-GAN in our previous work {[}35{]}, which employs a PatchGAN to refine \(x_{s}\) via adversarial learning, is also shown. As expected, \(x_{s}\) is blurred, while increasing residual depth progressively restores fine details, demonstrating the effectiveness of the decomposed latent design. Compared with the final HL-VAE outputs, VAE-GAN reconstructions are less realistic and physically consistent. This is because adversarial training relies on a discriminator's binary true/fake judgment, which can be unstable and insufficient, whereas RVQ more stably captures diverse residual patterns.

Quantitative evaluation is conducted using the commonly used Structural Similarity Index Measure (SSIM) and Peak Signal-to-Noise Ratio (PSNR). SSIM evaluates structural and perceptual similarity, while PSNR measures pixel-level reconstruction fidelity. Both metrics are computed on 500 testing images for HL-VAE at two residual depths and for VAE-GAN. The results show that HL-VAE consistently achieves higher SSIM and PSNR values, confirming its superior reconstruction quality and generalization capability.

\begin{figure}[!htbp]
\centering
\includegraphics[width=4.07143in,height=3.34526in]{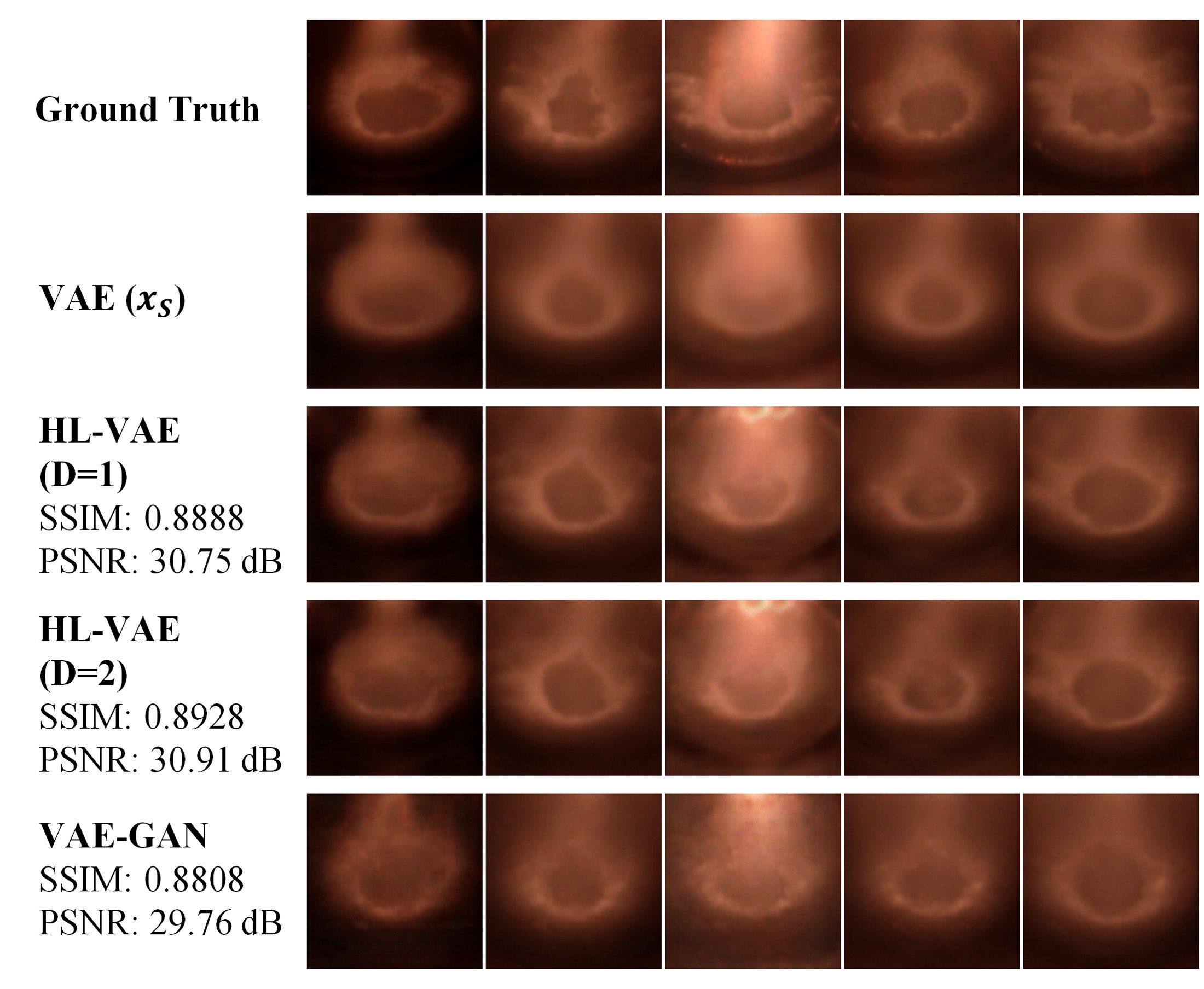}
\caption*{Figure 6. Weld Pool Reconstruction Comparison.}
\end{figure}

\hypertarget{temporal-modelling-of-latent-dynamics}{%
\subsection{Temporal Modelling of Latent Dynamics}\label{temporal-modelling-of-latent-dynamics}}

The objective of temporal modelling is to simulate the evolution of the latent representation \((\mu,M)\) under planned welding speed sequences \(u\), enabling predictive weld pool views for real-time human-robot collaboration. Since weld pool evolution is an inertial process influenced by both instantaneous speed and accumulated thermal history, latent dynamics must account for historical dependence.

The temporal evolution of the structural latent \(\mu\) is modeled using an LSTM network as a state transition function:

\[\begin{array}{r}
\mu_{\tau + 1} = f_{LSTM}\left( \mu_{\tau},u_{t} \right),\tau = t,t + 1,\cdots,T - 1\tag{6}
\end{array}\]

Recursive prediction is achieved by feeding each predicted latent state back as input for the next step, enabling long-horizon forecasts under arbitrary command sequences. To improve stability, the network is trained in a multi-step manner: instead of minimizing only one-step prediction error, the LSTM is unrolled across extended horizons and the accumulated error over multiple steps is optimized {[}32{]}.

To synthesize stochastic details, the discrete code map \(M\) is modeled as a conditional discrete distribution dependent on the predicted structural state:

\[\begin{array}{r}
M_{\tau + 1}\sim p\left( M \middle| \mu_{\tau + 1} \right)\tag{7}
\end{array}\]

For practical implementation, the continuous latent space is partitioned into \(N = 100\)clusters. For each cluster, the top three most frequent code indices at each spatial location \((h,w)\) are recorded and normalized to form categorical distributions. During inference, the nearest cluster to the predicted \(\mu_{\tau + 1}\) is selected, and indices are sampled to construct the residual map.

The predictive capability of the proposed HL-VAE framework was evaluated using both experimental data and simulation. Figure 7 shows the temporal evolution of the structural latent states \(\mu\) under both experimental and simulated welding speeds. First, the LSTM model iteratively updates its internal memory using the input and true latent states during the initialization phase, building a meaningful summary of historical information. Then, in the prediction phase, the model continues for 100 steps (10 s), recursively generating the next prediction from the input speed and the previous latent state. Under the experimental branch, the predicted latent components closely track the ground-truth latent states derived from the VAE encoder, demonstrating accurate dynamic modeling and long-horizon trend prediction. Under the simulated branch, the latent trajectories evolve consistently with the simulated speed input.

\begin{figure}[!htbp]
\centering
\includegraphics[width=5.81739in,height=4.56571in]{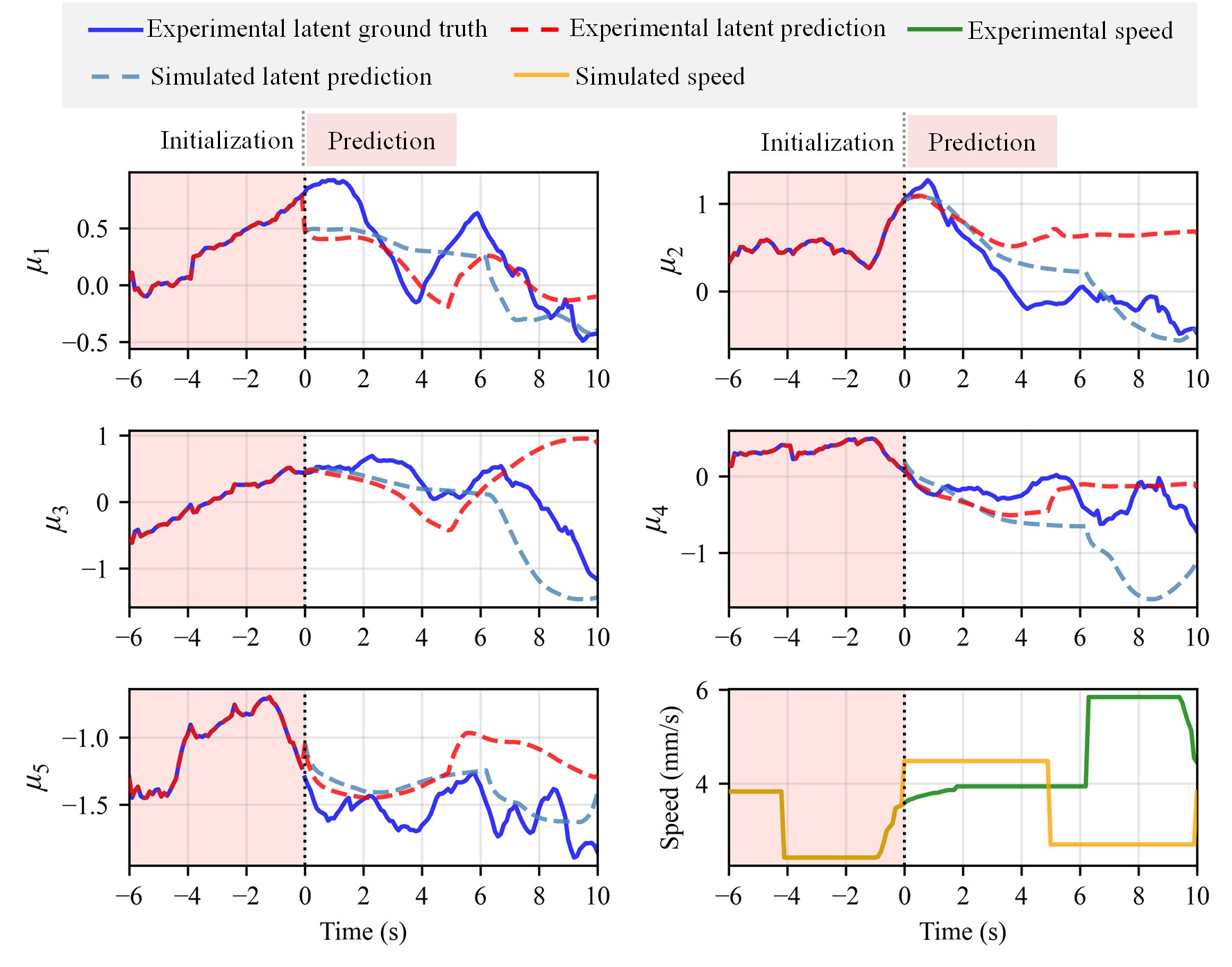}
\caption*{Figure 7. Long-term Latent State Prediction.}
\end{figure}

Figure 8 presents the corresponding weld pool image predictions for the prediction phase. In experimental verification, the predicted images closely match the ground-truth images, while the simulation results exhibit physically consistent evolution under unseen speed inputs. These results demonstrate that HL-VAE provides accurate predictive modeling and strong generalization capability.

Real-time weld pool prediction is achieved at the same frequency as image capture (10 Hz). The Python program runs on a high-performance CPU (Intel Core i7-13700K, 3.4 GHz, 16 cores). Since temporal prediction is propagated in a low-dimensional latent space \(\mu\), and stochastic details \(M\) are then added by lightweight sampling, the overall computational cost is negligible. The lightweight design of the HL-VAE enables weld pool reconstruction and refinement in 0.02 s per image, supporting synchronous 10 Hz generation and display of all weld pool views.

\begin{figure}[!htbp]
\centering
\includegraphics[width=6in,height=2.04653in]{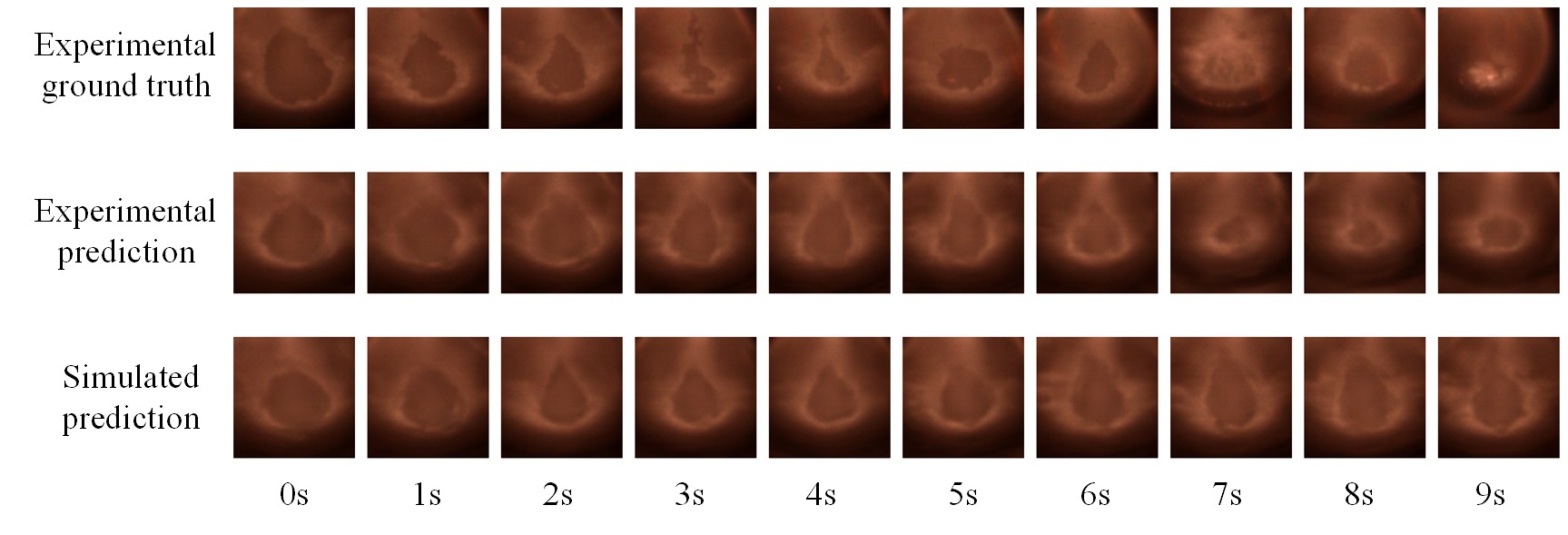}
\caption*{Figure 8. Long-term Weld Pool Prediction.}
\end{figure}

\section{Experiments}

With the predictive weld-pool visualization integrated into the HRC system, the operator shifts from reactive correction to proactive control. Because GTAW exhibits slow, inertial dynamics, relying only on instantaneous video feedback forces decisions to be made from lagged visual cues, which can lead to untimely corrections or overshoot. By contrast, a forward-looking weld-pool view lets the operator visually assess the effect of the current command, evaluate long-horizon error relative to the reference weld pool, and adjust speed proactively. We therefore conduct comparative experiments to verify the effectiveness of this assistance.

In experiments, disturbances were deliberately introduced by shifting the executed welding speed by ±1 mm/s, and the operator was required to counteract the disturbance and restore the weld pool to the reference state. Figure 9 illustrates the evaluation metrics, including the settling time \(t_{s}\), which evaluates response speed in regulating topside weld pool evolution, and the average backside bead width error \(e_{b}\), which measures post-welding penetration accuracy. The settling time \(t_{s}\) is defined as the time required for the pool width \(w\) to return to the target value (\(w^{*} = 3.5\text{\,}mm\).) after a disturbance, where \(w\) is extracted frame by frame using a weld pool segmentation model {[}36{]}. After welding, the backside bead is photographed with a physical scale for pixel-to-millimeter calibration, and the bead edge profile is manually extracted using Adobe Photoshop. The evenness metric \(e_{b}\) is defined as the Mean Absolute Deviation of the backside bead width \(w_{b}\) measured along the weld length, quantifying penetration uniformity. A lower \(e_{b}\) indicates more even penetration, reflecting more stable process control. Three trained operators (A, B, C) were evaluated under welding conditions with and without predictive views.

\begin{figure}[!htbp]
\centering
\includegraphics[width=3.56273in,height=2.41762in]{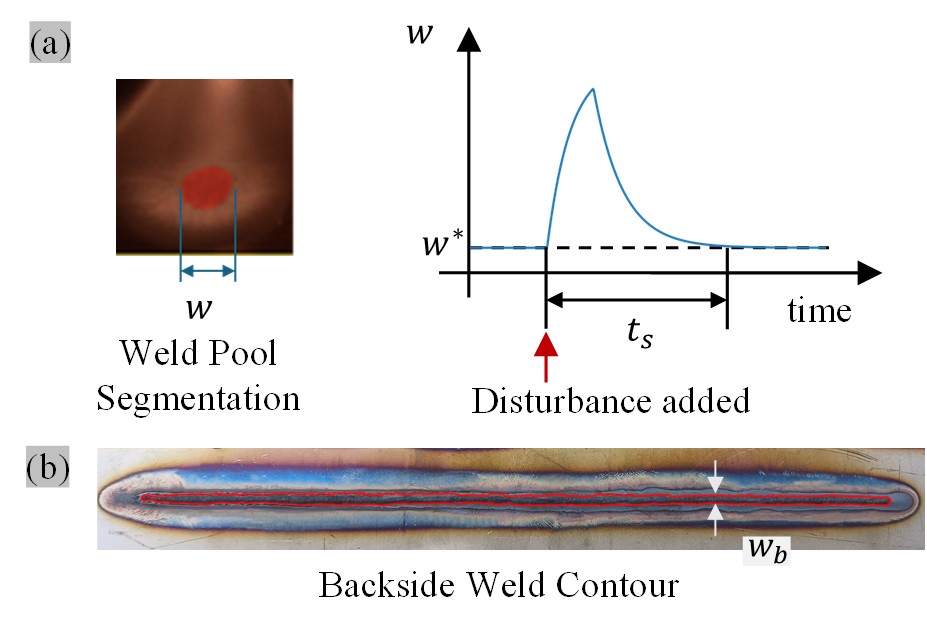}
\caption*{Figure 9. Experimental Metrics Illustration. (a) Setting time \(t_{s}\); (b) Backside bead width \(w_{b}\) measurement.}
\end{figure}

Figure 10 exemplifies the baseline condition without predictive views. A speed disturbance was introduced by shifting the executed welding speed by −1 mm/s, and the operator was required to counteract the disturbance and regulate the weld pool back to the reference state. Because the effect of each adjustment is not immediately visible, the operator adopted conservative, incremental speed changes based on delayed visual feedback. As a result, the weld pool width error decayed slowly after reaching its peak following the disturbance. The settling time \(t_{s}\) in the baseline condition was 2.75 s. As shown in the figure, the backside bead width exhibited noticeable unevenness (\(e_{b} = 0.49\) mm), caused by delayed adjustment and inaccuracies inherent in purely reactive control.

\begin{figure}[!htbp]
\centering
\includegraphics[width=4.76508in,height=3.45854in]{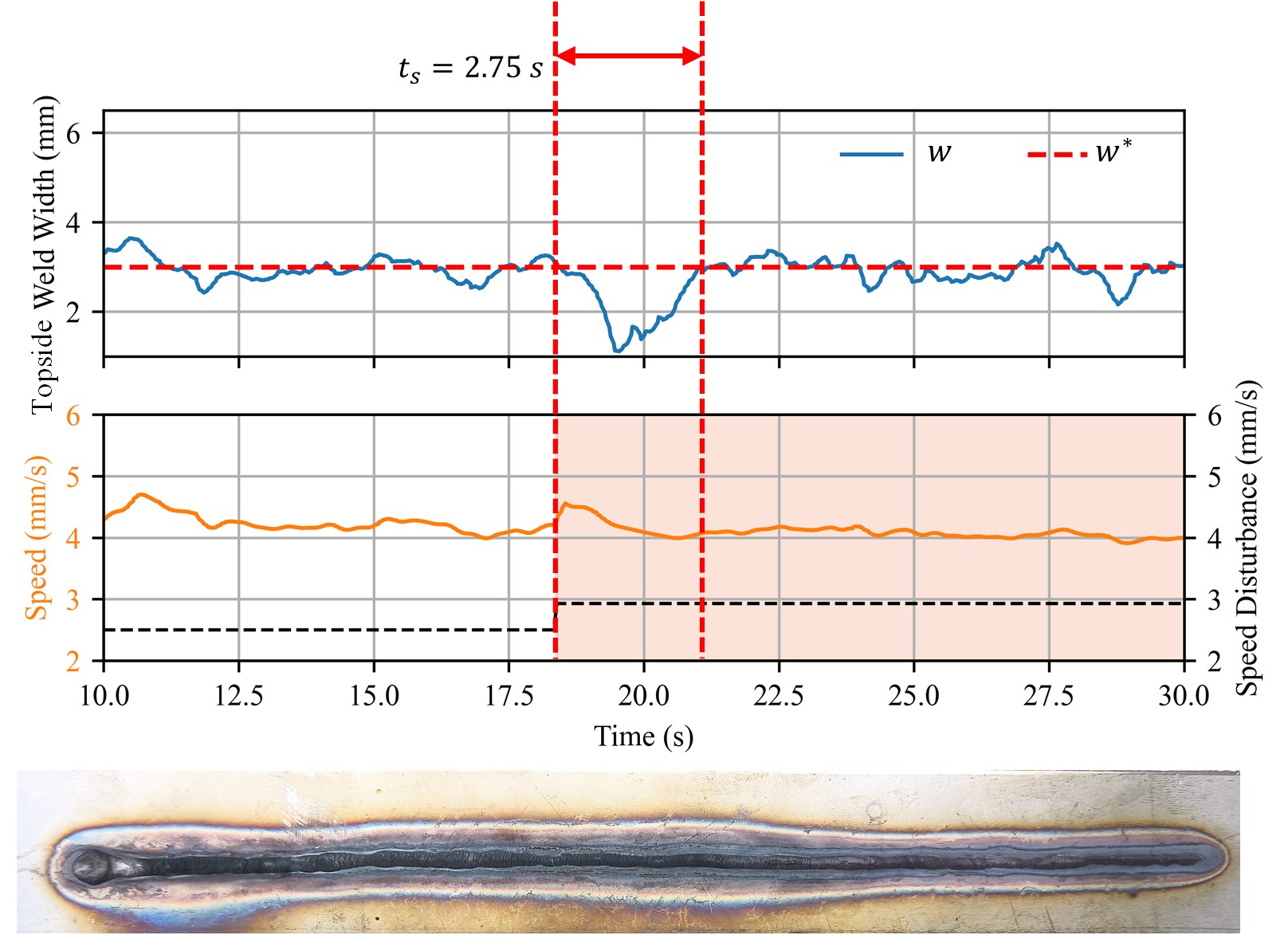}
\caption*{Figure 10. Human Control Experiment without Predictive Views (Operator A).}
\end{figure}

Figure 11 shows the case with predictive views enabled. With the generative model deployed, the 5 s predicted view explicitly reveals the steady-state consequence of the current speed command. After the disturbance (+1 mm/s speed shift in execution), the operator issued a swift correction by lowering the welding speed, followed only by minor fine-tuning. This is because the preview provided by the generative model turns decision-making from relying on delayed process feedback to directly comparing the predicted steady-state error with the reference. As a result, control of the inertial welding process becomes an instant adjustment aimed at eliminating static error. This leads to a shorter settling time, measured as \(t_{s} = 1.47\ s\) , which is clearly shorter than the baseline. The backside weld was also more even and smooth (\(e_{b} = 0.38\) mm). This demonstrates that predictive weld-pool visualization significantly enhances GTAW control by enabling a proactive control mode.

\begin{figure}[!htbp]
\centering
\includegraphics[width=4.84149in,height=3.43331in]{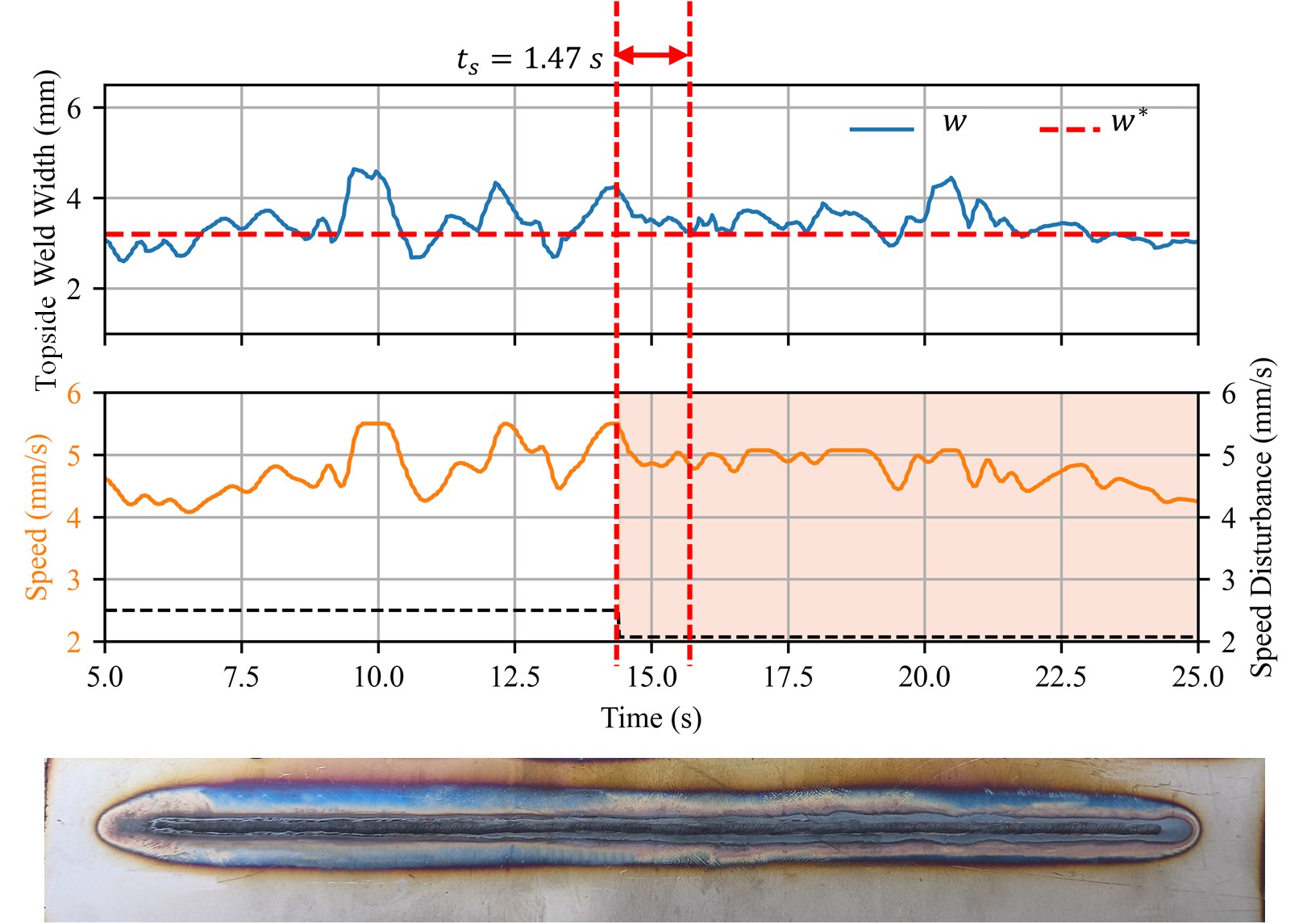}
\caption*{Figure 11. Human Control Experiment with Predictive Views (Operator A).}
\end{figure}

The experimental results from three operators are summarized in Table 1. Despite individual differences, the transition from reactive to proactive control consistently improves performance, as verified by reductions in both \(t_{s}\) and \(e_{b}\).

\begin{table}[!htbp]
\centering
\caption*{Table 1. Performance comparison of human control}
\begin{tabular}[]{@{}
  >{\raggedright\arraybackslash}p{(\linewidth - 6\tabcolsep) * \real{0.2022}}
  >{\raggedright\arraybackslash}p{(\linewidth - 6\tabcolsep) * \real{0.3494}}
  >{\raggedright\arraybackslash}p{(\linewidth - 6\tabcolsep) * \real{0.1983}}
  >{\raggedright\arraybackslash}p{(\linewidth - 6\tabcolsep) * \real{0.2501}}@{}}
\toprule
\begin{minipage}[b]{\linewidth}\raggedright
Operator
\end{minipage} & \begin{minipage}[b]{\linewidth}\raggedright
Predictive View
\end{minipage} & \begin{minipage}[b]{\linewidth}\raggedright
\(t_{s}\)/s
\end{minipage} & \begin{minipage}[b]{\linewidth}\raggedright
\(e_{b}\)/mm
\end{minipage} \\
\midrule

\bottomrule

A & Y & 1.47 & 0.38 \\
B & Y & 1.62 & 0.41 \\
C & Y & 1.55 & 0.33 \\
A & N & 2.75 & 0.49 \\
B & N & 2.98 & 0.53 \\
C & N & 2.88 & 0.55 \\
\end{tabular}
\end{table}

\section{Conclusions}

This paper has presented an intelligent human--robot collaboration (HRC) system for adaptive control of Gas Tungsten Arc Welding (GTAW). The system enables operators, including those with limited training, to conduct adaptive welding control remotely, where robotic execution compensates for the physical limitations of manual operation and embedded generative AI reduces the cognitive demands of interpreting dynamic weld pool behavior.

First, the robotic system overcomes the physical restrictions of human operators. A collaborative robot equipped with a welding torch executes operator commands captured through the handlers, while a VR-based digital twin provides an immersive, multi-source information display. This integration physically separates operators from the harsh welding environment, enables clearer observation through integrated sensing and visualization, and lowers the skill threshold for directly manipulating the welding torch. Through synchronized digital replicas and intuitive VR control, operators can focus on high-level decision-making rather than low-level execution.

Then, a generative model AI model predicts and visually renders future weld pool states in the digital twin. Specifically, a Hybrid Latent Variational Autoencoder (HL-VAE) is designed to decouple the structural and stochastic image content of the weld pool within the latent space. Subsequent temporal modeling of these latent dynamics, leveraging recurrent neural architectures, forecasts their evolution under specific process inputs and reconstructs them into visually interpretable predictions. This predictive capability supports proactive decision-making, allowing operators to anticipate the consequences of their actions and achieve more stable control. Human experiments verify the effectiveness of integrating robotic execution, human intelligence, and generative AI.

Despite the success of this work, effectively representing imaged or even multi-modal welding data in a low-dimensional manifold remains challenging due to the diverse and noisy nature of real-world scenarios. Future work will continue in this direction, exploring a more principled partitioning of structural and stochastic information within a rigorous information-theoretic framework. In addition, we will investigate dynamic modeling and control for arc welding based on high-quality human demonstrations, leveraging more robust latent representations.

\textbf{Declaration}: This manuscript benefited from editorial refinement using ChatGPT (OpenAI), under the authors' direction and with full review and approval by the authors. The tool was used to improve clarity and readability; the authors take full responsibility for the content.

\textbf{References}

\begin{enumerate}
\def\labelenumi{\arabic{enumi}.}
\item
  ``World Robotics - Industrial Robots.'' International Federation of Robotics. https://ifr.org/wr-industrial-robots/ (accessed Jan. 21, 2025).
\item
  Y. M. Zhang, Y. P. Yang, W. Zhang, and S. J. Na, ``Advanced Welding Manufacturing: A Brief Analysis and Review of Challenges and Solutions,'' J. Manuf. Sci. Eng. Trans. ASME, 2020, doi: 10.1115/1.4047947.
\item
  Zhang, Y., Wang, Q., \& Liu, Y. (2021). Adaptive intelligent welding manufacturing. Weld J, 100(1), 63-83.
\item
  Doumanidis, C.C. and Hardt, D.E., 1991. Multivariable adaptive control of thermal properties during welding.
\item
  R. Yu, Y. Cao, J. Martin, O. Chiang, and Y. M. Zhang, ``Robotizing double-electrode GMAW process through learning from human welders,'' J. Manuf. Process., 2024, doi: 10.1016/j.jmapro.2023.12.008.
\item
  Cao, Y., Zhou, Q., Yuan, W., Ye, Q., Popa, D., \& Zhang, Y. (2024). Human-robot collaborative assembly and welding: A review and analysis of the state of the art.~Journal of Manufacturing Processes,~131, 1388-1403.
\item
  Gualtieri, L., Rauch, E., \& Vidoni, R. (2021). Emerging research fields in safety and ergonomics in industrial collaborative robotics: A systematic literature review.~Robotics and Computer-Integrated Manufacturing,~67, 101998.
\item
  ``World Robotics - Industrial Robots.'' International Federation of Robotics. https://ifr.org/wr-industrial-robots/ (accessed Jan. 21, 2025).
\item
  Y. M. Zhang, Y. P. Yang, W. Zhang, and S. J. Na, ``Advanced Welding Manufacturing: A Brief Analysis and Review of Challenges and Solutions,'' J. Manuf. Sci. Eng. Trans. ASME, 2020, doi: 10.1115/1.4047947.
\item
  Zhang, Y., Wang, Q., \& Liu, Y. (2021). Adaptive intelligent welding manufacturing. Weld J, 100(1), 63-83.
\item
  Lee, D., \& Han, K. (2024). Vision-based construction robot for real-time automated welding with human-robot interaction.~Automation in Construction,~168, 105782.
\item
  Doumanidis, C.C. and Hardt, D.E., 1991. Multivariable adaptive control of thermal properties during welding.
\item
  R. Yu, Y. Cao, J. Martin, O. Chiang, and Y. M. Zhang, ``Robotizing double-electrode GMAW process through learning from human welders,'' J. Manuf. Process., 2024, doi: 10.1016/j.jmapro.2023.12.008.
\item
  Ebel, L. C., Zuther, P., Maass, J., \& Sheikhi, S. (2020). Motion signal processing for a remote gas metal arc welding application.~Robotics,~9(2), 30.
\item
  Su, Y. P., Chen, X. Q., Zhou, T., Pretty, C., \& Chase, G. (2021). Mixed reality-enhanced intuitive teleoperation with hybrid virtual fixtures for intelligent robotic welding.~Applied Sciences,~11(23), 11280.
\item
  Cao, Y., \& Zhang, Y. (2025). Control of DE-GMAW through human--robot collaboration.~Welding in the World,~69(5), 1459-1468.
\item
  Cao, Y., Chen, H., \& Zhang, Y. (2025). Monitoring of DE-GMAW process in human--robot collaboration.~Welding in the World,~69(5), 1427-1435.
\item
  Niu, Y., Wu, S., Cheng, F., \& Wang, Z. (2025). A Weld Pool Morphology Acquisition and Visualization System Based on an In Situ Calibrated Analytical Solution and Virtual Reality.~Sensors,~25(9), 2711.
\item
  Chatzidakis, S., Giuliano, D., Slade, J., Tang, W., Miller, R., Reeves, S., ... \& Howard, R. (2021). A Versatile Remediation Module for Remote Repair of Spent Nuclear Fuel and High-Level Waste Storage Containers.~Nuclear Technology,~207(5), 750-760.
\item
  Liu, Y., \& Zhang, Y. (2017). Fusing machine algorithm with welder intelligence for adaptive welding robots.~Journal of Manufacturing Processes,~27, 18-25.
\item
  Cao, Y., Guo, S., \& Zhang, Y. (2025). Robotizing GTAW through learning human response.~Welding in the World, 1-11.
\item
  Zhou, J., Mohammad, A., Zeng, T., Axinte, D., Wright, I., \& March, R. (2025). Teaching robots to weld by leveraging human expertise.~Robotics and Computer-Integrated Manufacturing,~95, 103027.
\item
  Y. Ye, T. Zhou, and J. Du, ``Robot-Assisted Immersive Kinematic Experience Transfer for Welding Training,'' J. Comput. Civ. Eng., 2023, doi: 10.1061/jccee5.cpeng-5138.
\item
  Zhang, Z., Ikeura, R., Hayakawa, S., \& Wang, Z. (2024, August). Reinforcement learning-based skill acquisition study for linearly heated robots. In~2024 IEEE International Conference on Mechatronics and Automation (ICMA)~(pp. 1397-1402). IEEE.
\item
  Wu, H., Zhai, X., Zheng, H., Liao, Z., Xu, Z., \& Zhou, X. (2025). Learning stability-guaranteed skill and adaptive control strategies from demonstrations for heterogeneous component robotic machining.~Journal of Manufacturing Processes,~151, 506-520.
\item
  R. Yu, Y. Cao, H. Chen, Q. Ye, and Y. M. Zhang, ``Deep learning based real-time and in-situ monitoring of weld penetration: Where we are and what are needed revolutionary solutions?,'' J. Manuf. Process., 93(2023): 15-46. doi: 10.1016/j.jmapro.2023.03.011.
\item
  S. Zou, Z. Wang, S. Hu, W. Wang, and Y. Cao, ``Control of weld penetration depth using relative fluctuation coefficient as feedback,'' J. Intell. Manuf., 2020, doi: 10.1007/s10845-019-01506-8.
\item
  Y. Cao, Z. Wang, S. Hu, and T. Wang, ``Adaptive Predictive Control of Backside Weld Width in Pulsed Gas Metal Arc Welding Using Electrical Characteristic Signals as Feedback,'' IEEE Trans. Control Syst. Technol., 2023, doi: 10.1109/TCST.2023.3258064.
\item
  P. Wang, J. Kershaw, M. Russell, J. Zhang, Y. Zhang, and R. X. Gao, ``Data-driven process characterization and adaptive control in robotic arc welding,'' CIRP Ann., 2022, doi: 10.1016/j.cirp.2022.04.046.
\item
  Wang, Q., Jiao, W., \& Zhang, Y. (2020). Deep learning-empowered digital twin for visualized weld joint growth monitoring and penetration control.~Journal of Manufacturing Systems,~57, 429-439.
\item
  Wang, Q., Jiao, W., Wang, P., \& Zhang, Y. (2020). Digital twin for human-robot interactive welding and welder behavior analysis.~IEEE/CAA Journal of Automatica Sinica,~8(2), 334-343.
\item
  Zhang, W. J., Liu, Y. K., Wang, X., \& Zhang, Y. M. (2012). Characterization of three-dimensional weld pool surface in GTAW. Welding journal, 91(7), 195s-203s.
\item
  A. Arora, G. Roy, T. DebRoy, Unusual wavy weld pool boundary from dimensional analysis, Scripta Materialia 60 (2) (2009) 68--71. {[}38{]}
\item
  Y. Deng, L. Shu, Y. Wang, D. Ma, Z. Luo, P. Jiang, A data-driven framework for molten pool morphology prediction in laser welding with dimensionless numbers discovery, Journal of Manufacturing Processes 151 (2025) 826--842.
\item
  Y. Cao, Q. Ye, Y. M. Zhang, ``Synthesizing Weld Pool Dynamics via VAE-GAN to Enhance Human Control Performance,'' In review for J. Manuf. Process., 2025
\item
  Yu, R., Kershaw, J., Wang, P., \& Zhang, Y. (2021). Real-time recognition of arc weld pool using image segmentation network. Journal of Manufacturing Processes, 72, 159-167.
\item
  Lee, D., Kim, C., Kim, S., Cho, M. and Han, W.S., 2022. Autoregressive image generation using residual quantization. In~Proceedings of the IEEE/CVF conference on computer vision and pattern recognition~(pp. 11523-11532).
\end{enumerate}

\end{document}